\documentclass[authoryear,preprint,review,12pt]{elsarticle}

\usepackage{amssymb}
\usepackage{amsmath}
\usepackage{booktabs}
\usepackage{multirow}
\usepackage{color}
\usepackage{natbib}
\usepackage[table,xcdraw]{xcolor}

\journal{Nuclear Physics B}

\begin{document}

\begin{frontmatter}



\title{Underwater Variable Zoom: Depth-Guided Perception Network for Underwater Image Enhancement}

\author[1]{Zhixiong~Huang}
\ead{hzxcyanwind@mail.dlut.edu.cn}

\author[1]{Xinying~Wang}
\ead{wangxinying@mail.dlut.edu.cn}

\author[2]{Chengpei~Xu}
\ead{Chengpei.Xu@unsw.edu.au}

\author[3]{Jinjiang~Li}
\ead{lijinjiang@sdtbu.edu.cn}

\author[4,5]{Lin~Feng\corref{cor1}}
\ead{fenglin@dlut.edu.cn}
\cortext[cor1]{Corresponding author}

\affiliation[1]{organization={School of Computer Science and Technology, Dalian University of Technology},
	city={Dalian 116081},
	state={Liaoning},
	country={China}}    
\affiliation[2]{organization={MIoT \& IPIN Lab, School of Minerals and Energy Resources Engineering, University of New South Wales},
	city={Sydney 2052},
	state={NSW},
	country={Australia}}             
\affiliation[3]{organization={School of Computer Science and Technology, Shandong Technology and Business University},
	city={Yantai 264005},
	state={Shandong},
	country={China}}
\affiliation[4]{organization={School of Information and Communication Engineering, Dalian Minzu University},
	city={Dalian 116600},
	state={Liaoning},
	country={China}}
\affiliation[5]{organization={School of Innovation Experiment, Dalian University of Technology},
	city={Dalian 116081},
	state={Liaoning},
	country={China}}

\begin{abstract}
	The underwater images often suffer from color deviations and blurred details. To address these issues, many methods employ networks with an encoder/decoder structure to enhance the images. However, the direct skip connection overlooks the differences between pre- and post-features, and deep network learning introduces information loss. This paper presents an underwater image enhancement network that focuses on pre-post differences. The network utilizes a multi-scale input and output framework to facilitate the underwater image enhancement process. A novel cross-wise transformer module (CTM) is introduced to guide the interactive learning of features from different periods, thereby enhancing the emphasis on detail-degraded regions. To compensate for the information loss within the deep network, a feature supplement module (FSM) is devised for each learning stage. FSM merges the multi-scale input features, effectively enhancing the visibility of underwater images. Experimental results across several datasets demonstrate that the integrated modules yield significant enhancements in network performance. The proposed network exhibits outstanding performance in both visual comparisons and quantitative metrics. Furthermore, the network also exhibits good adaptability in additional visual tasks without the need for parameter tuning. The code and results are released in https://github.com/WindySprint/UVZ.
\end{abstract}

%


\end{frontmatter}


\section{Introduction}
As a crucial information carrier, underwater images play an important role in ocean tasks\citep{1}. Unfortunately, underwater images often encounter various degradation problems due to two main factors \citep{2}. Firstly, different wavelengths of light decay at varying rates in the underwater medium, resulting in a prevalent blue-green color deviation. Secondly, the scattering effect of suspended particles hinders normal light propagation, leading to problems such as low contrast and blurred details \citep{3}. Meanwhile, these degradations worsen with the increasing distance from the imaging scene to the camera \citep{4}. Therefore, underwater image enhancement (UIE) methods are widely used in visual interpretation disciplines, such as biology, ecology, and animal behavior, aiding researchers in conducting subsequent analysis. Meanwhile, as a pre-processing step for computer vision tasks such as image segmentation, saliency detection, and moving object detection, UIE methods can significantly improve the accuracy and robustness of algorithms, providing strong support for ocean monitoring, ecological protection, and resource exploration.

\begin{figure*}
	\centering
	\includegraphics[width=12cm]{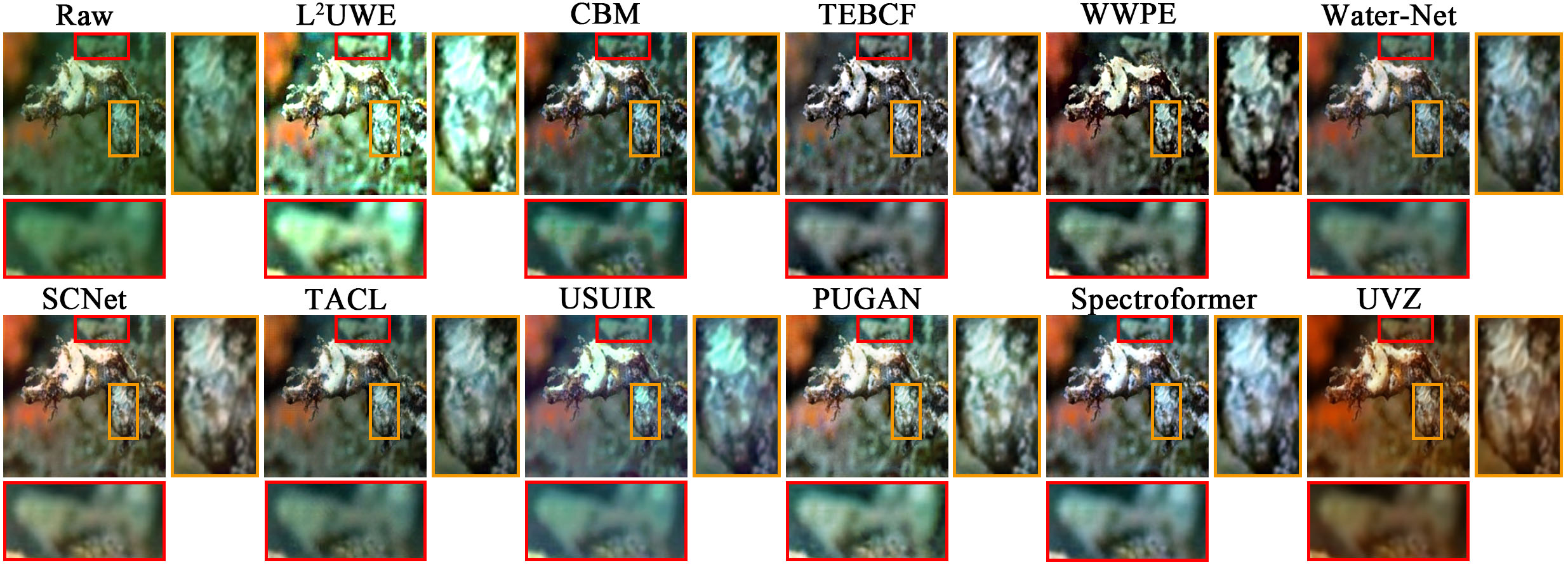}
	\caption{Near-far scenarios enhancement comparison by different methods, where the orange box represents the enlarged near scene and the red box represents the enlarged far scene. The top row shows the raw image, the results of L\(^2\)UWE \protect\citep{5}, CBM \protect\citep{6}, TEBCF \protect\citep{7}, WWPE \protect\citep{8}, and Water-Net \protect\citep{9}. The bottom row shows the results of SCNet \protect\citep{10}, TACL \protect\citep{11}, USUIR \protect\citep{12}, PUGAN \protect\citep{13}, Spectroformer \protect\citep{14}, and our UVZ.}
	\label{Fig:1}
\end{figure*}

\begin{figure*}
	\centering
	\includegraphics[width=12cm]{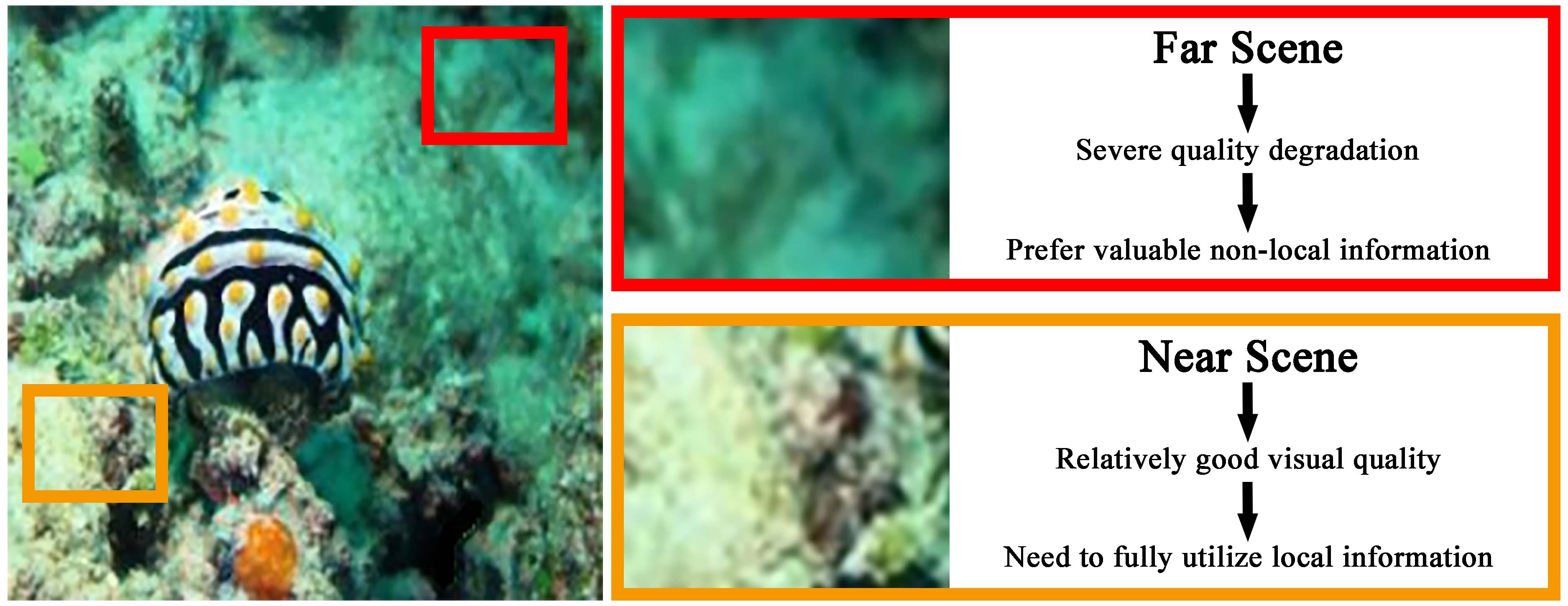}
	\caption{Motivation of our proposed depth-guided perception. The degradation degree varies significantly between the near scene and far scene. In comparison, the visual quality of the former and the adjacent regions exhibits relatively better performance. For the two scenes, we should implement different feature learning strategies.}
	\label{Fig:2}
\end{figure*}

In the early days of  UIE field, researchers have proposed model-free and model-based traditional methods. Model-free methods \citep{5,15,16} enhances underwater images by modifying pixels in degraded channels and blurred regions. On the other hand, model-based methods \citep{7,17} introduce physical cues to build the underwater degradation process, and reverse it to restore underwater images. In recent years, deep learning methods \citep{11,18,19} have shown remarkable enhancements, leveraging benchmark data and miscellaneous networks. Despite significant progress achieved with the above methods, challenges emerge when addressing widely varying degrees of regional degradation. As shown in Figure~\ref{Fig:1}, most UIE methods have considerable visual improvement in the near scene, but still retain notable color deviation in the far scene. Figure~\ref{Fig:2} highlights the significant differences between near and far scenes, thereby underscoring the motivation behind this work. A more detailed description is provided in Section~\ref{subsection:3.1}.

For more precise color enhancement, we believe that the integration of depth information related to scene distance is essential, which benefits the region division in underwater images. Previous works \citep{12,13,20} attempted to incorporate depth information related to scene distance, contributing to the response of degraded regions. However, direct weight adjustments may restrict the exploration of visual cues, as the adjacent features of far and near points exhibit varying gains in the enhancement process. In near regions of relatively good quality, local information from surrounding features is critical to the quality of the enhanced image, whereas in the far regions of heavily degradation, non-local information from other regions can provide more valuable insights. With the above analysis, we aim to construct an adaptive region enhancement method in underwater scenes, achieving comprehensive visual enhancement by differentiating in utilizing local and non-local information. To achieve this goal, there are two main challenges:(1) On the one hand, the local feature capture capability helps to enhance the near region where the surrounding information is enough. On the other hand, the remote dependency establishment capability helps to enhance the far region where the surrounding information is lost. However, due to feature space differences, there is mutual interference between these two capabilities. (2) Diverse underwater scenes tremendously raise the difficulty of predicting the depth map.

Specifically, we propose a novel UIE method based on depth-guided perception, UVZ, which includes depth estimation network (DEN), auxiliary supervision network (ASN), and depth-guided enhancement network (DGEN). Firstly, the proposed model generates a depth map by DEN, and uses ASN to further supervise the rationality of the depth map during training. When enhancing each point, the depth information is used to determine the contribution of non-local and local information. Meanwhile, the desired features for different regions are captured by non-local and local branches in DGEN. To adapt to varying receptive fields, we design an R\(^3\)S transformation for the estimated depth map, guiding feature fusion in different regions. Through the above process, UVZ effectively and comprehensively enhances the color, contrast, and clarity of underwater images.

The main contributions of this paper are as follows:

• We design a novel depth-guided perception framework, UVZ, for non-homogeneous degraded underwater scenes, which utilizes adapted depth maps to execute a learning strategy that integrates non-local and local features.

• To improve the distinction of degraded regions, we incorporate an attention estimation-scene reconstruction training process, aiming to strengthen the connection between depth maps and underwater scenes.

• Compared to state-of-the-art methods, UVZ exhibits more attractive results across extensive qualitative and quantitative comparisons, and demonstrates generalization abilities to other visual tasks without parameter tuning.

\section{Related work}
\subsection{Traditional Methods}
In response to the degradation characteristics of images, model-free methods notably enhance the contrast and saturation of the images. The early studies focused on adjusting image pixel values in the spatial domain. Ancuti et al. \citep{15} fused color compensated and white balance images to obtain natural underwater image. Marques et al. \citep{5} used contrast to guide the generation of images with salient details and brightness enhancement. As the research progressed, some methods introduced morphology and structural properties to understand underwater images. Based on the concept of contour bougie morphology, Yuan et al. \citep{6} enhanced the image quality by mitigating scattering blur and contrast stretching. Kang et al. \citep{21} decompose the image into mean intensity, contrast, and structure, the three components are fused in different ways to obtain the enhancement result. Additionally, Zhang et al. \citep{3} introduced the attenuation characteristics to correct the color, and introduced the integral and squared integral maps to enhance the contrast. Overall, model-free methods overlook the characteristics of underwater degradation process, especially the depth between the object and the camera \citep{22,23}, leading to color distortion or improper enhancement.

Model-based methods explore the process of underwater image degradation and use priors and assumptions to model this process. Alenezi et al. \citep{17} introduced gradient and absorption differences associated with color channels into background light and transmission map estimation. Zhou et al. \citep{23} introduced light attenuation features and feature prior in estimating background light and transmission map. Chandrasekar et al. \citep{24} combine underwater image formation models and depth information to remove backscatter, and utilize lightweight networks and retinex model to further enhance the image. In addition, some methods introduced prior knowledge from other domains to improve visual quality. Berman et al. \citep{25} used the haze-lines model to enhance the attenuated colors. Yuan et al. \citep{7} enhanced the visibility of images by dark channel prior and color compensation in two color spaces, respectively. Due to the variability of underwater environments, the parameters and priors set by such methods are limited and cannot fully adapt to specific scenarios.

\subsection{Deep Learning Methods}
In recent years, deep learning methods have made significant progress in UIE tasks. Classical methods were devoted to optimize the network structure to achieve better visual enhancement. Jiang et al. \citep{26} incorporated the laplacian pyramid to construct a lightweight multi-scale network, and improved the performance by recursive strategy and residual learning. Zhang et al. \citep{27} introduced multi-level sub-networks to enrich the feature representation, and proposed a triple attention module to selectively cascade different features. To enrich the input features, some studies have introduced other color spaces. Wang et al. \citep{28} introduced HSV space for the first time to adjust the brightness and saturation of underwater images. Zhou et al. \citep{18} combined RGB and HSV color spaces to enhance the underwater images, and used attention and color-guided map to guide the network to perceive critical information. Furthermore, the researchers have delved into the structural composition of images from diverse perspectives of data distribution. By observing from the perspective of color spectrum distribution, Rao et al. \citep{19} proposed a two-stage network using texture cues and color cues to enhance the underwater images. Wang et al. \citep{29} integrated the wavelet transform and color compensation mechanism in the network to perform multi-level enhancement of details and colors.

To avoid the need for paired underwater images, unsupervised learning methods have received considerable attention. Among them, generative adversarial network (GAN) has been a prevalent framework. Jiang et al. \citep{30} employed global-local adversarial mechanism to supervise image realism. Li et al. \citep{31} introduced comparative learning into UIE and utilized the discriminator to evaluate the quality of different regions. Beyond GAN, other learning methods such as transfer learning and comparative learning also show immense potential. Jiang et al. \citep{32} migrated image dehazing from air to the underwater domain, thus avoiding the reliance of paired underwater images. Liu et al. \citep{11} constructed a twin adversarial contrastive learning framework and utilized underwater object detection tasks to guide image enhancement. 

Due to the lack of degradation cues, relying solely on network mapping may not comprehensively enhance underwater images, resulting in residual blue-green deviations. To tackle this challenge, we introduced depth information associated with the degradation process to guide the network.

\subsection{Depth-related Methods}
The widespread application of depth information in image dehazing \citep{33,34,35} has enlightened the understanding of underwater scenes in the UIE field. Traditional methods incorporated depth estimation to mitigate underwater degradation effects on image quality. Zhou et al. \citep{36} used channel intensity prior and unsupervised method to generate two depth maps, further eliminated the blurring problem caused by scattering effects. Hou et al. \citep{37} improved the estimation of transmission map by the local background light and reflectance map, which can eliminate the influence of non-uniform light in underwater scenes. Simultaneously, deep learning methods utilized depth information to guide the scene weights for targeted distribution. Chen et al. \citep{20} introduced position attention and dilated convolution into the depth estimation network, guiding weight allocation by the predicted depth map. Fu et al. \citep{12} reconstructed the raw image by predicting a depth-related transmission map, thereby obtaining the scene radiance for degradation. Cong et al. \citep{13} used the depth map and the physical model for image restoration, with dual-discriminators constraining the generation of the restored image and the depth map. 

The above method incorporates depth information to adapt degradation scenes, but low-quality features still impact the enhancement of distant regions. Considering this, we employ adaptive learning for regions at various distances, aiming to establish differentiated region enhancement.

\section{Methodology}
\subsection{Overall Architecture}\label{subsection:3.1}
Due to the regional degradation gap in underwater scenes, local dependence for perceiving surrounding features against remote dependence for perceiving global structures, it is nontrivial to learn the two perceptual capabilities from the diverse samples. The existing UIE methods eliminate most color deviation, but certain regions may still be affected by blue-green, particularly in far-filed regions with severe degradation. Because the enhancement process of far point is mainly dominated by the surrounding degraded features, and the local information instead hinders the underwater color correction. The remote dependence considers the overall scene from a macroscopic perspective, but lacks a precise definition and identification of the degree of regional degradation. To locate regions with severe degradation, we note the correlation between depth and degradation in underwater imaging models. A reasonable depth map can provide a intuitive assessment of the image quality in different regions, which is important for fusing local and global perceptual capabilities. In this work, we focus on building an integrated UIE framework guided by physical cues, aiming to achieve comprehensive color enhancement for diverse underwater degraded scenes.

\begin{figure*}
	\centering
	\includegraphics[width=12cm]{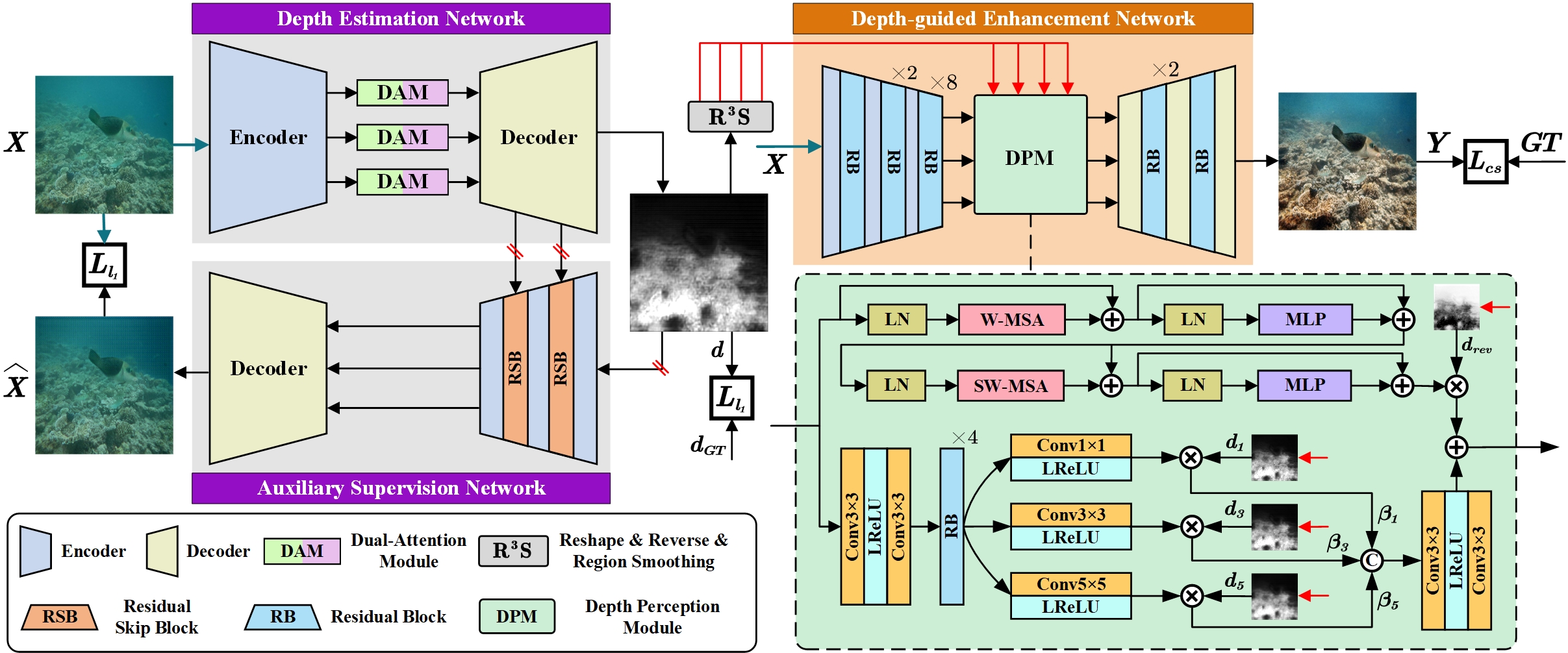}
	\caption{The two-stage framework of UVZ, including DEN, DGEN, and ASN. All sub-networks adopt a standard encoder/decoder architecture, where the red slash indicates that ASN is only used for the training. In the first stage, for a raw image \(X\), DEN and ASN generate the depth map \(d\) and the regression image \(\widehat{X}\), respectively. For the second stage, with inputs \(X\) and \(d\), DGEN generates the enhanced image \(Y\).}
	\label{Fig:3}
\end{figure*}

As shown in Figure~\ref{Fig:3}, UVZ constitutes a two-stage UIE framework. In the first stage, we input the raw image \(X\in \mathbb{R}^{H\times W\times 3}\) to the depth estimation network (DEN), and output the depth map \(d\in \mathbb{R}^{H\times W\times 1}\) through encoding-decoding and two scale changes. Within the process of feature mapping, a dual-attention module (DAM) is introduced to monitor severely degraded regions from channel and spatial perspectives, allowing for a more explicit definition of the near and far regions. By introducing scene reconstruction and cyclic consistency loss, CycleGAN \citep{38} ensures consistency between the target images and the source images. Inspired by this, we introduce the auxiliary supervision network (ASN) to generate the regression map \(\widehat{X}\in \mathbb{R}^{H\times W\times 3}\) in the training phase, and jointly supervise the losses of DEN and ASN to promote greater correlation between \(d\) and \(X\). To refine the restoration of \(\widehat{X}\), ASN fuses the decoded features from DEN via residual skip blocks (RSB), which further promotes the rationality of \(\widehat{X}\) on the overall scene.

In the second stage, we design a depth perception module (DPM) for near and far feature learning. DPM consists of two distinct branches for non-local and local feature learning, with selective fusion guided by \(d\) from the previous stage. In response to the DPM, we design an R\(^3\)S transformation for \(d\), as shown in Figure~\ref{Fig:4}. Using the adjusted depth maps and the learnable parameters, the depth-guided enhancement network (DGEN) constructs a fusion strategy with receptive field adaptation to combine non-local and local features more rationally. Meanwhile, DGEN incorporates multiple residual blocks (RB) to solve the gradient vanishing. Both RB and RSB extract features through convolution-activation-convolution structures and residual connection, while the latter includes an additional 1 × 1 convolution block for fusion.

\subsection{Depth Perception Module}
\begin{figure*}[h!]
	\centering
	\includegraphics[width=7.5cm]{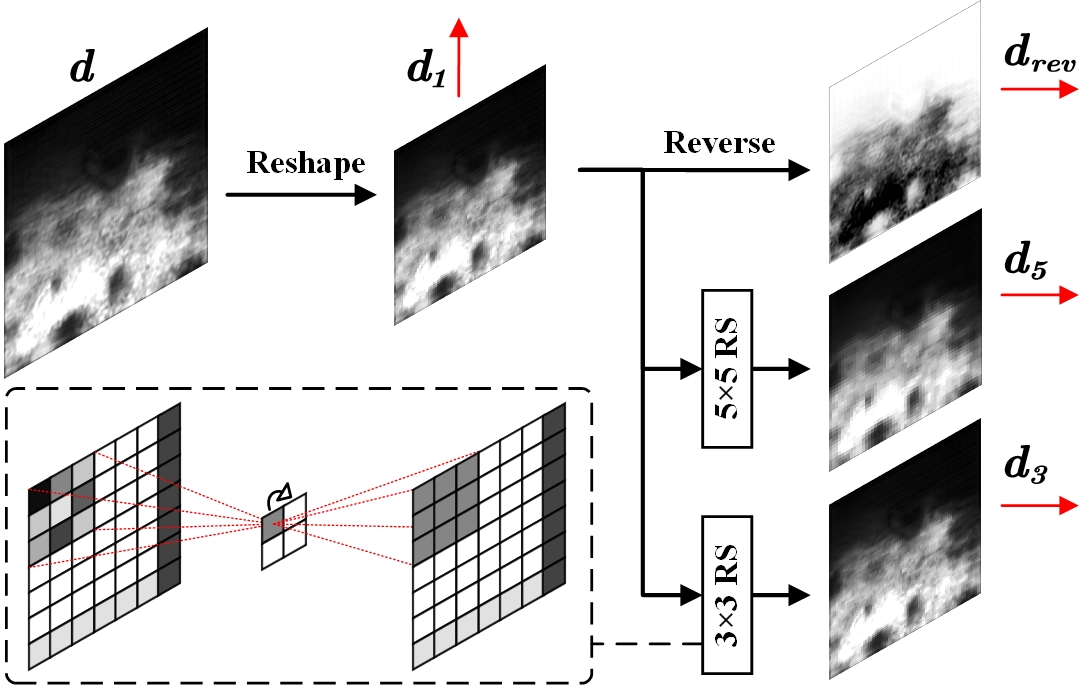}
	\caption{R\(^3\)S transformation of a depth map. The input is the depth map d and the outputs are the feature maps \(d_{rev}\), \(d_1\), \(d_3\), and \(d_5\). The process within the dashed line is a 3 × 3 region smoothing (RS).}
	\label{Fig:4}
\end{figure*}

We constructed a DPM that guides perception with depth, and adaptively enhance different regions through non-local and local learning branches. Specifically, to adapt the receptive fields of the two branches, the DPM applies an R\(^3\)S transformation to the depth map \(d\). As shown in Figure~\ref{Fig:4}, we first reshape d to \(d\) to \(d_1\in \mathbb{R}^{h\times w\times 1}\) consistent with the encoded feature scale, which is performed by nearest interpolation. Second, we alter the weight distribution of the far and near scenes by reversing the depth map \(d_{rev}=1-d\), so that the non-local branch focuses more on high-quality information. In addition, we introduce a region smoothing (RS) process to remove noise from degraded regions, thereby avoiding the interference of far points with higher degradation on local information capture. RS divides the regions based on the size of the convolution kernel in the local branch, calculating the mean value of each region for normalization. The specific process is as follows:

\vspace{-1cm}
\begin{align}
d_1=R_1 \cup R_2 \cup \ldots \cup R_n, \nonumber\\
\mu_{i}=\frac{1}{k^2} \sum_{x_{i j} \in R_i} x_{i j},
\end{align}
where \(R_1,R_2,\ldots,R_n\) represent the different division regions, \(n=(h×w)/k^2\). \(x_{ij}\) represents the jth pixel of region \(R_i\), and all pixels in that region are represented by the mean value \(\mu_{i}\). Ultimately, we employ RS to obtain \(d_3\) and \(d_5\) corresponding to 3 × 3 and 5 × 5 convolutions. By guiding the selective fusion in DPM, the transformed depth map will enable the network to develop a reasonable perception of the degraded regions, making the enhanced colors more natural.

We adopt the Swin Transformer \citep{39} to implement non-local branch for capturing dependencies. Compared to the Vision Transformer, Swin Transformer captures remote dependencies at diverse scales by constructing hierarchical feature mappings, which allows it to capture details of different shapes (e.g., fish, plants, and stones, etc.) in underwater scenes. Additionally, Swin Transformer introduces a shifted windows mechanism to enhance the receptive field, further extending the ability to facilitate non-local feature interactions. The input encoded features are first partitioned into a non-overlapping patch sequence \(f_i=[p_1,p_2,\ldots,p_m]\), then flattened into linear vectors and computed as follows:

\vspace{-1cm}
\begin{align}
\hat{f}_i=MSA\left(LN \left(f_i\right)\right)+f_i,  \nonumber\\
f_i=MLP\left(LN \left(\hat{f}_i\right)\right)+\hat{f}_i,
\end{align}
where \(LN(\cdot)\) represents layer normalization, and \(MLP(\cdot)\) represents multi-layer perceptron. \(MSA(\cdot)\) represents multi-head self-attention, which alternates between window-based and shift-window-based  partitioning strategies. We merge the extracted patch sequences to obtain the feature map \(f_t\), and multiply it with \(d_{rev}\) to obtain the non-local result \(f_r=f_t\times d_{rev}\). By non-local branch, we reduce the attention to surrounding information in far regions, avoiding the poor-quality features dominating the enhancement process. Instead, high-quality features will serve as a more favorable option for the region.

On the other hand, local information should play a crucial role in near regions with good imaging. Compared to the non-local region, adjacent features tend to be more helpful for the enhancement process in that region. In the local branch, we propose a local feature extractor based on multi-kernel convolutions, including four 3 × 3 convolutions, four residual blocks, and three convolutions with different kernel sizes (1, 3, 5), where the activation functions are LeakyReLU. To improve the spatial distribution perception of underwater scenes, multi-kernel convolutions are applied to obtain more perceptual fields, thereby enhancing details of varying sizes in multi-scale encoding features. We multiply the convolutional features \(f_{c1}\),\(f_{c3}\),\(f_{c5}\) with the corresponding depth maps \(d_1\), \(d_3\), \(d_5\), and dynamically adjust the weights by three learnable parameters \(\beta_1\), \(\beta_3\), and \(\beta_5\):
\begin{equation}
f_{c k}=d_k \times f_{c k} \times \beta_k, k=1,3,5. \\
\end{equation}

We concatenate the three outputs on the channel dimension and generate local branch results by composite convolutional blocks. Combining the results of non-local and local branch to output the following feature \(f_o \in \mathbb{R}^{h\times w\times c}\):

\vspace{-0.5cm}
\begin{equation}
f_o=f_r+cat\left(f_{c 1}, f_{c 3}, f_{c 5}\right)_{c l c}, \\
\end{equation}
where \(cat(\cdot)\) represents the concatenate operation on channel dimension, and \((\cdot)_{clc}\) represents the convolution-activation-convolution structure.

\subsection{Dual-Attention Module}
\begin{figure}
	\centering
	\includegraphics[width=12cm]{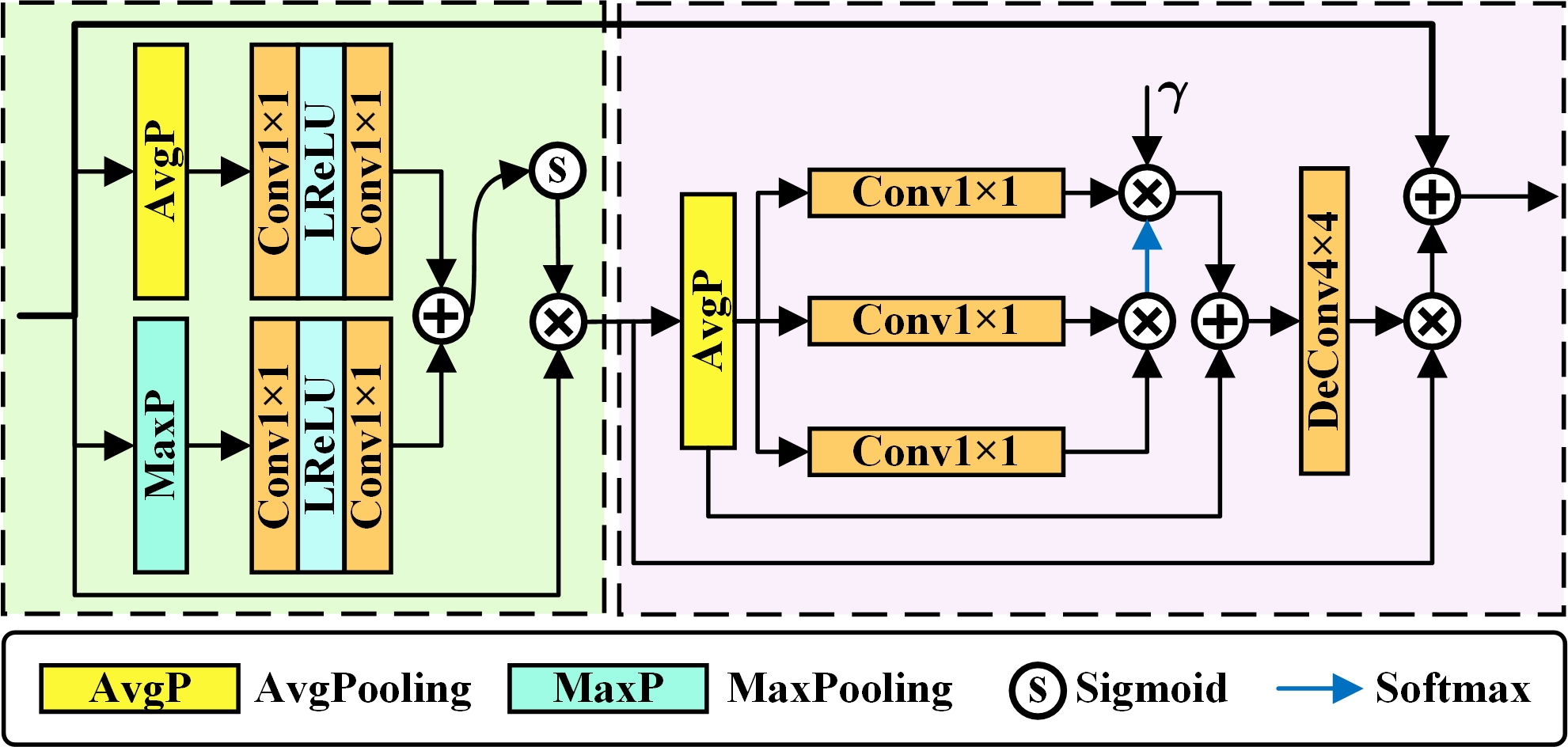}
	\caption{Structure of the Dual-Attention Module (DAM)}
	\label{Fig:5}
\end{figure}

In complex underwater imaging processes, the depth-related degradation attenuation is quite common. As shown in Figure~\ref{Fig:5}, we design a DAM that includes channel and spatial monitoring mechanisms to generate reasonable depth maps. First, DAM compresses the encoded feature \(f\) into two channel descriptors through different pooling operations. Then, the channel dependence used to adapt the feature representation is captured by the squeeze-excitation process \citep{40}. The above process can be represented as follows:

\vspace{-0.5cm}
\begin{equation}
f_{c}=f \times\left(A v g P(f)_{c l c}+M a x P(f)_{c l c}\right)_{s i g}, \\
\end{equation}
where \(AvgP(\cdot)\) and \(MaxP(\cdot)\) represent the average pooling layer and maximum pooling layer, respectively, and \((\cdot)_{sig}\) is the sigmoid function. The features annotated with channel weights \(f_c\) is input into the next stage, allowing regions with different degradation degrees to have distinct weights. First, we use the average pooling layer to reduce the feature scale, and then capture the spatial dependence by the query vector \(q\), the key vector \(k\), and the value vector \(v\). In this process, we introduce a learnable parameter \(\gamma\) to dynamically adjust the learning intensity:

\vspace{-1cm}
\begin{align}
q=k=v=\left(A v g P(f_{c})\right)_c, \nonumber \\
f_{s}=f_{c}+\gamma \times\left((\mathrm{q} \times k)_{soft} \times v\right),
\end{align}
where \((\cdot)_{soft}\) is a softmax function, and \(\gamma\) is a learnable parameter used to adjust the intensity. To label spatial weights \(f_s\) on input features, we first use a deconvolution \((\cdot)_{dc}\) to reshape \(f_s\) into input scale. Subsequently, we fuse the two attention features and perform residual connections with the original input \(f\). As a result, the proposed model can capture the hidden feature \(f_h\) with regional degradation prominence:

\vspace{-0.5cm}
\begin{equation}
f_h=f+f_{c} \times\left(f_{s}\right)_{d c}.
\end{equation}

Finally, we use the decoder to generate corresponding depth maps, guiding the network to perceive different degraded regions. In various underwater scenarios, although the region division of the depth map may not be precise enough, extensive experiments have demonstrated that the UVZ has good generalization and generates results with natural colors.

\subsection{Training Strategy}
The UVZ training process is supervised in two stages. In the first stage, we compute losses for the outputs of DEN and ASN, respectively, which is an offline pre-training process. During the second stage, we use DEN with fixed parameters to assist DGEN in generating the enhanced image, thereby obtaining relevant losses. The first stage of training is conducted by synthesized dataset \citep{41}, which contains underwater images and corresponding depth maps of five open ocean waters and coastal waters. We randomly selected 600 image pairs to train DEN and ASN for 100 epochs, and calculated the following \(l_1\) loss:
\begin{equation}
L_{l_1}=\frac{\lambda_1}{h w} \sum|d-d^{G T}| + \frac{1}{c h w} \sum|\widehat{X}-X|,
\end{equation}
where \(d\) and \(d^{GT}\) represent the estimated depth map and the ground truth (GT), \(\widehat{X}\) and \(X \) represent the regression image and the underwater image. We set the hyperparameter \(\lambda_1=3\) to balance the two losses. By adding the regression process in the training, \(d\) and \(X\) will be more closely structured.

In the second stage, we selected 1200 image pairs from the CYCLE \citep{42} dataset, with 1080 pairs used for 100 epochs training and the remaining 120 pairs used for subsequent testing. To ensure the enhanced images with pleasing colors and details, we formulate the total loss \(L_s\) of DGEN using charbonnier loss and ssim loss. The former measures pixel similarity, while the latter evaluates structural similarity. The formula is as follows:
\begin{equation}
L_s=\sqrt{|| x-\left.y\right||^2+\varepsilon^2}+\lambda_2(1-\operatorname{SSIM}(x, y)),
\end{equation}
where \(x\) and \(y\) are the enhanced image and GT. \(\lambda_2\) and \(\varepsilon\) are the hyperparameters for balancing losses, which were set to 0.5 and 10\(^{-3}\), respectively.

\section{Experiment}
\subsection{Implementation Details}
To assess the performance of UVZ, we use the remaining 120 image pairs from CYCLE dataset for comparison experiments. Additionally, we used four benchmark datasets: LNRUD \citep{43}, LSUI \citep{44}, UFO \citep{45}, and UIEB \citep{9}. We randomly selected 500 images from LNRUD, 400 images from LSUI, 120 images from UFO, and 80 images from UIEB for more comprehensive testing.

The proposed model was trained and tested on NVIDIA GeForce RTX 2080 Ti GPUs using the PyTorch framework. We initialized the learning rate to 0.0002 and adjusted it gradually with Adam optimizer. For the CYCLE dataset, we measure the similarity between the enhanced results and the reference images by three common referenced metrics (i.e., PSNR, SSIM and MSE). For all datasets, we evaluate the performance by three no-reference metrics (i.e., UIQM \citep{46}, UICM and NIQE).

\subsection{Comparison on reference dataset}
Our UVZ was compared with some state-of-the-art methods, including four traditional methods (i.e., L\(^2\)UWE \protect\citep{5}, CBM \protect\citep{6}, TEBCF \protect\citep{7}, WWPE \protect\citep{8}), four deep learning methods (i.e., Water-Net \protect\citep{9}, SCNet \protect\citep{10}, TACL \protect\citep{11}, Spectroformer \protect\citep{14}), and two depth-related methods (i.e., USUIR \protect\citep{12}, PUGAN \protect\citep{13}).

\begin{figure*}[h!]
	\centering
	\includegraphics[width=12cm]{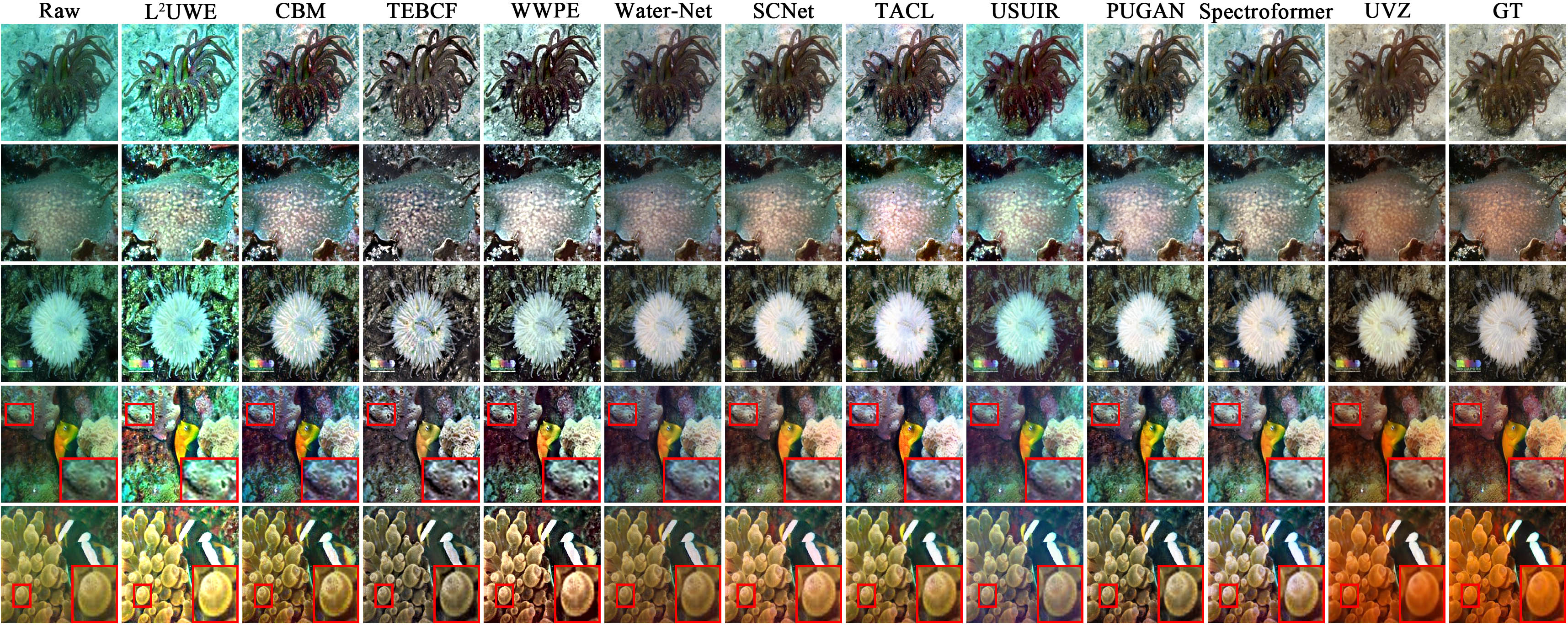}
	\caption{Enhancement comparison on the CYCLE dataset. From left to right are the raw images, the results of L\(^2\)UWE \protect\citep{5}, CBM \protect\citep{6}, TEBCF \protect\citep{7}, WWPE \protect\citep{8}, and Water-Net \protect\citep{9}, SCNet \protect\citep{10}, TACL \protect\citep{11}, USUIR \protect\citep{12}, PUGAN \protect\citep{13}, Spectroformer \protect\citep{14}, the proposed UVZ, and the ground truth. The red box represents the enlarged details.}
	\label{Fig:6}
\end{figure*}

\begin{figure*}[h!]
	\centering
	\includegraphics[width=12cm]{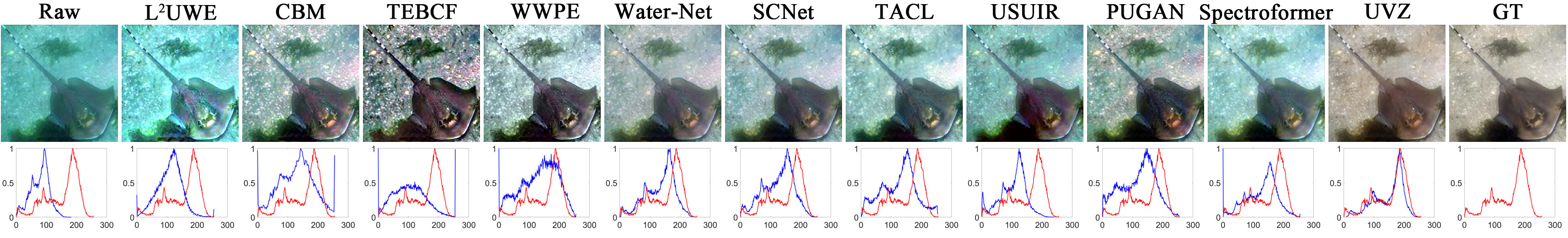}
	\caption{Histogram comparison of red channel on the CYCLE dataset, with the raw image, the enhanced images from all methods, and GT on top, and the corresponding red channel histogram on bottom. The red and blue lines are the data distributions of the GT and enhanced images.}
	\label{Fig:7}
\end{figure*}

\begin{figure*}[h!]
	\centering
	\setlength{\belowcaptionskip}{-0.3cm}
	\includegraphics[width=12cm]{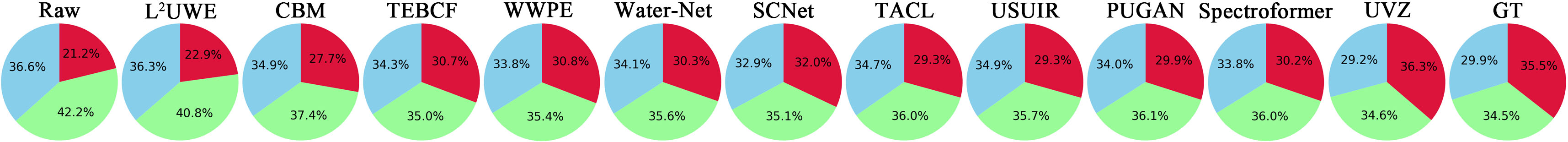}
	\caption{Distribution comparison of RGB three-color on the CYCLE dataset}
	\label{Fig:8}
\end{figure*}

Figure~\ref{Fig:6} shows the enhancement results of all methods on the CYCLE dataset. L\(^2\)UWE and USUIR improve image contrast, but noticeable color deviations persist. CBM and Water-Net have corrected colors to some extent, but the enlarged image details are blurred. TEBCF significantly enhances the details of the image but results in darkened brightness and color confusion. WWPE, TACL, PUGAN, and Spectraformer alleviate the effects of color deviation, but enlarged details are still affected by blue-green. SCNet and our UVZ effectively eliminate color deviation and enhance image visibility, with the latter highlighting more vibrant colors. Moreover, we also exhibit the histograms of the red channel in Figure~\ref{Fig:7}, which is the most severely lost during underwater imaging. The curves of UVZ and GT overlap the most, indicating the closest similarity. To further observe the color distribution, we show the pixel statistics on the RGB channels in Figure~\ref{Fig:8}. All methods enhance the degraded red channel to some extent, with UVZ being the most prominent. Its distribution on the red, green, and blue channels is 36.3\%, 36.1\%, and 29.2\% respectively, which is more consistent with GT's 35.5\%, 34.5\%, and 29.9\%.

\begin{table*}[h!]
	\tiny
	\centering
	\renewcommand{\arraystretch}{1.5}
	\setlength\tabcolsep{5pt}
	\caption{Metrics comparison of all methods on the CYCLE dataset (\textbf{optimal}, \underline{sub-optimal})}
	\begin{tabular}{cccccccc}
		\toprule
		& \textbf{PSNR↑} & \textbf{SSIM↑} & \textbf{MSE↓}    & \textbf{UIQM↑} & \textbf{UICM↑} & \textbf{NIQE↑} & \textbf{Runtime(s)} \\ \midrule
		L\(^2\)UWE   {\citep{5}}        & 14.55 & 0.6440 & 0.06195 & 2.634 & -76.35 & 5.045 & 3.6725     \\
		CBM {\citep{6}}            & 21.48 & 0.7461 & 0.01745 & 4.151 & -31.83 &  \underline{6.548} & 0.2425     \\
		TEBCF {\citep{7}}          & 18.74 & 0.6839 & 0.02090 & 4.606 & -13.71 & 5.137 & 1.6340     \\
		WWPE {\citep{8}}           & 17.40 & 0.6412 & 0.02560 & 4.291 & -16.63 & 6.387 & 0.3207     \\ \hline
		Water-Net {\citep{9}}      &  \underline{23.23} &  \underline{0.8196} & 0.01150 & 4.573 & -16.43 & 5.310 & 0.9613     \\
		SCNet {\citep{10}}         & 22.48 & 0.8116 &  \underline{0.01081} &  \underline{4.629} &  \underline{-11.01} & 5.203 & 0.0520     \\
		TACL {\citep{11}}          & 21.78 & 0.7701 & 0.01597 & 4.444 & -23.29 & 4.325 & 0.0595     \\
		Spectroformer {\citep{14}} & 21.42 & 0.7910 & 0.01639 & 4.332 & -20.74 & 5.017 &  \underline{0.0394}     \\ \hline
		USUIR {\citep{12}}         & 20.55 & 0.7739 & 0.02211 & 4.321 & -21.15 & 4.941 & \textbf{0.0255}     \\
		PUGAN {\citep{13}}         & 21.43 & 0.7790 & 0.01571 & 4.522 & -20.74 & 5.253 & 0.0488     \\ \hline		
		UVZ                    & \textbf{26.00} & \textbf{0.8721} & \textbf{0.00435} & \textbf{5.044} & \textbf{4.781}  & \textbf{6.586} & 0.0583   \\ \bottomrule   
	\end{tabular}
	\label{Tab:1}
\end{table*}

Table~\ref{Tab:1} shows the quantitative comparison on the CYCLE dataset, where UVZ outperforms other state-of-the-art methods in all quality assessments, particularly with 11.9\% and 8.9\% performance improvements in PSNR and UIQM. Compared to other methods, Water-Net is more effective in removing color deviation, while SCNet notably enhances the red channel. Therefore, these methods obtain most sub-optimal metrics in Table~\ref{Tab:1}. However, in terms of runtime, our method does not exhibit outstanding speed. To accurately guide region perception, UVZ adopts a two-stage design and multiple attentions. While these configurations require substantial computing resources, the model performance is nonetheless impressive.

\subsection{Comparison on no-reference dataset}
\begin{figure*}[h!]
	\centering
	\setlength{\belowcaptionskip}{-0.3cm}
	\includegraphics[width=12cm]{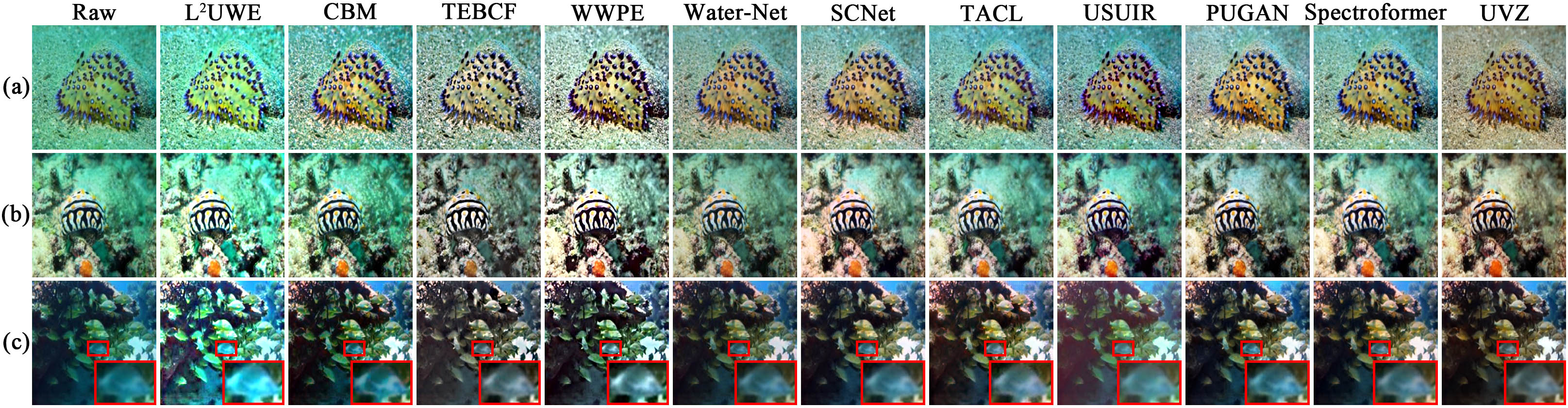}
	\caption{Enhancement comparison of all methods on the LNRUD dataset}
	\label{Fig:9}
\end{figure*}

\begin{figure*}[h!]
	\centering
	\setlength{\belowcaptionskip}{-0.3cm}
	\includegraphics[width=12cm]{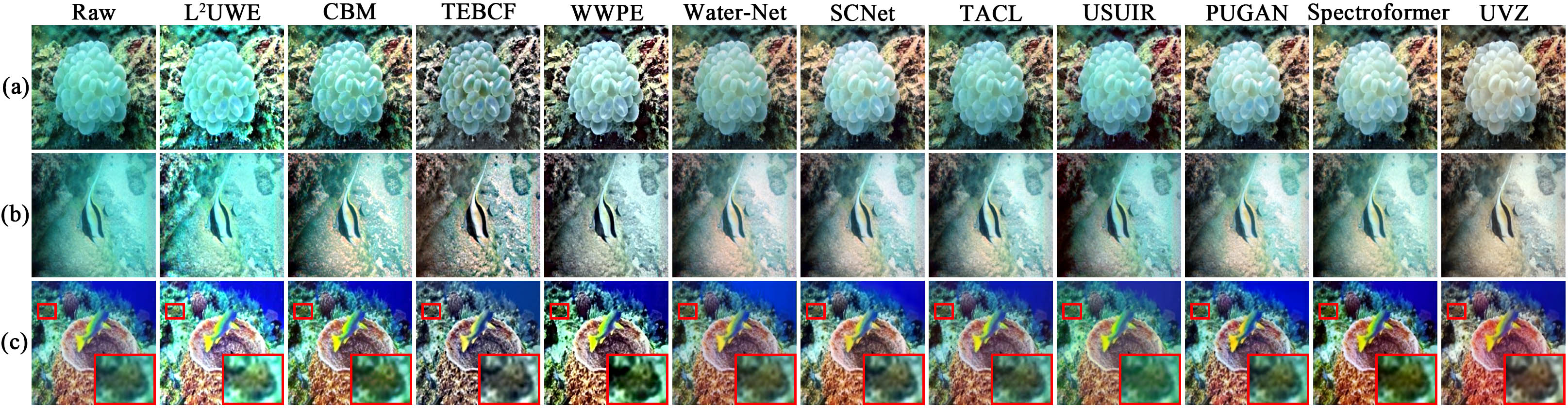}
	\caption{Enhancement comparison of all methods on the LSUI dataset}
	\label{Fig:10}
\end{figure*}

We present enhancement examples from LNRUD, LSUI, UFO and UIEB datasets in Figs.~9-12. In Figure~\ref{Fig:9} and Figure~\ref{Fig:10}, the color enhancement from the UVZ is the most pronounced and well-established. While other methods mitigate the effects of color deviation, blue-green color still dominates the results. We enlarged the fishes in the near-far scenes in Figure~\ref{Fig:9} (c) and Figure~\ref{Fig:10} (c), respectively. It can see that only TEBCF and UVZ effectively remove the blue-green color, but the dull color of the former reduces the attractiveness.

\begin{figure*}[h!]
	\centering
	\setlength{\belowcaptionskip}{-0.3cm}
	\includegraphics[width=12cm]{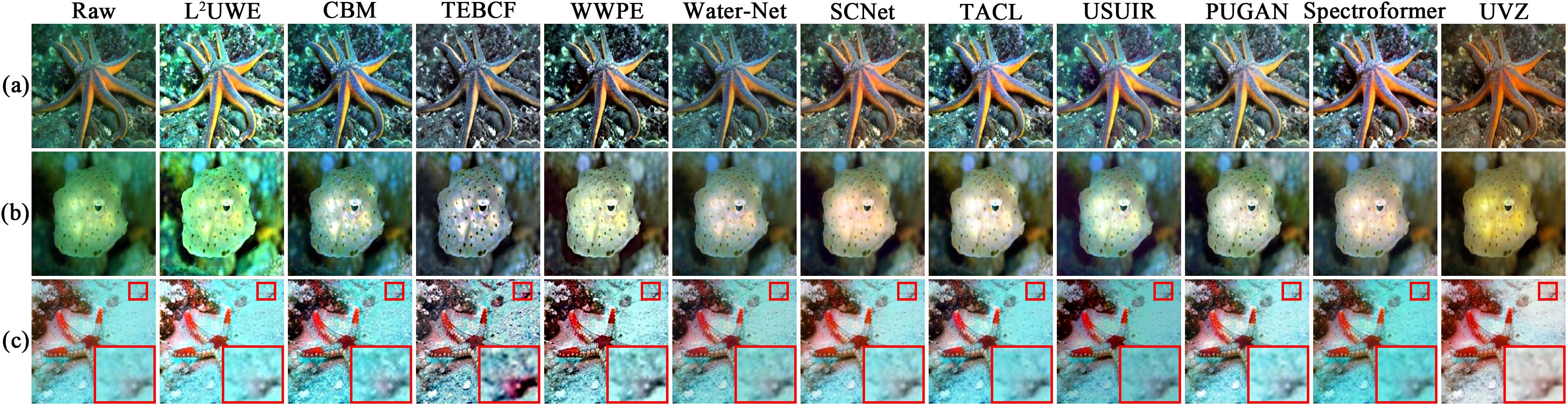}
	\caption{Enhancement comparison of all methods on the UFO dataset}
	\label{Fig:11}
\end{figure*}

\begin{figure*}[h!]
	\centering
	\setlength{\belowcaptionskip}{-0.3cm}
	\includegraphics[width=12cm]{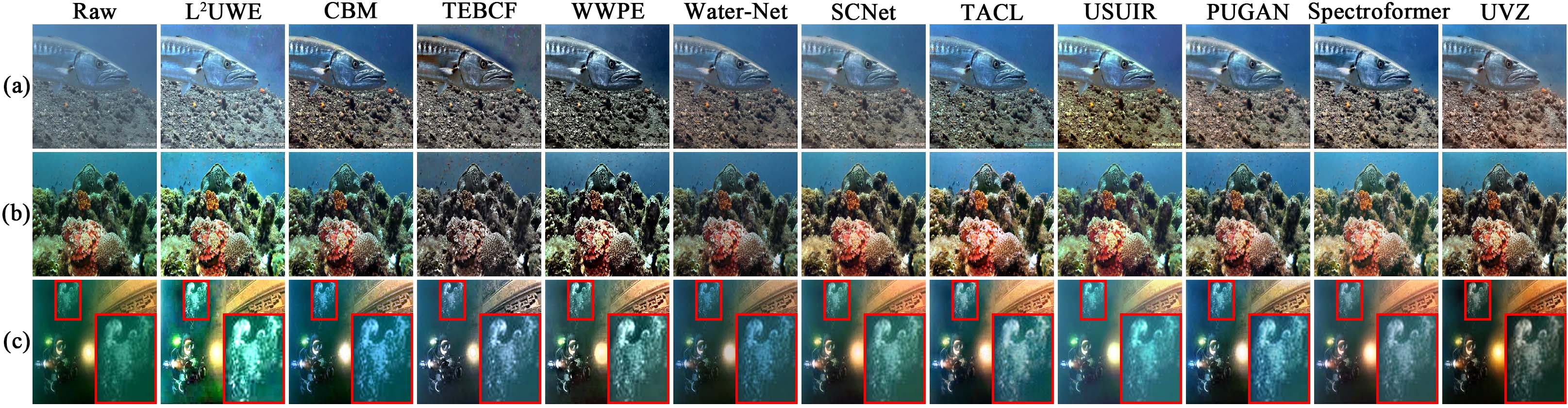}
	\caption{Enhancement comparison of all methods on the UIEB dataset}
	\label{Fig:12}
\end{figure*}

From Figure~\ref{Fig:11} and Figure~\ref{Fig:12}, it can be observed that L\(^2\)UWE exhibits bright contrast enhancement, but over-exposure affects the color. The enhanced results of TEBCF, WWPE and Water-Net have low brightness and poor clarity. CBM effectively enhances image details, while TACL, USUIR and PUGAN alleviate the impact of color deviation. However, blue-green colors still dominate in their enhanced images. Despite the significant saturation increase in SCNet and Spectroformer, they still show deviated colors in the far scenes. Benefiting from its depth perception ability, our UVZ has excellent performance in terms of color, saturation, and visibility, whether in near or far scenes.

\begin{table*}[h!]
	\tiny
	\centering
	\renewcommand{\arraystretch}{1.5}
	\setlength\tabcolsep{0.5pt}
	\caption{Metrics comparison of all methods on four benchmark datasets (\textbf{optimal}, \underline{sub-optimal})}
	\begin{tabular}{c
			>{\columncolor[HTML]{E8E8E8}}c 
			>{\columncolor[HTML]{E8E8E8}}c 
			>{\columncolor[HTML]{E8E8E8}}c ccc
			>{\columncolor[HTML]{E8E8E8}}c 
			>{\columncolor[HTML]{E8E8E8}}c 
			>{\columncolor[HTML]{E8E8E8}}c ccc}
		\toprule
		& \multicolumn{3}{c}{\cellcolor[HTML]{E8E8E8}LNRUD} & \multicolumn{3}{c}{LSUI} & \multicolumn{3}{c}{\cellcolor[HTML]{E8E8E8}UFO} & \multicolumn{3}{c}{UIEB} \\
		\multirow{-2}{*}{}                                & UIQM↑  & UICM↑   & NIQE↑  & UIQM↑  & UICM↑   & NIQE↑ & UIQM↑  & UICM↑  & NIQE↑ & UIQM↑  & UICM↑   & NIQE↑ \\ \midrule
		L\(^2\)UWE {\citep{5}}            & 2.688  & -74.80  & 5.322  & 2.882  & -72.28  & 5.220 & 2.800  & -68.97 & 4.385 & 2.716  & -73.48  & 6.702 \\
		CBM {\citep{6}}              & 4.037  & -31.29  & \textbf{6.684}  & 4.215  & -29.06  & \textbf{6.429} & 3.849  & -34.93 & 4.398 & 3.911  & -32.56  & \textbf{8.807} \\
		TEBCF {\citep{7}}              & 4.436  & \underline{-16.69}  & 5.485  & 4.635  & -14.98  & 5.340 & 4.223  & -20.31 & 4.808 & 4.240  & \underline{-16.75}  & 6.977 \\
		WWPE  {\citep{8}}              & 4.296  & -17.15  & 5.550  & 4.439  & -13.64  & 5.637 & 4.325  & -19.00 & 4.575 & 4.505  & -17.43  & \underline{7.494} \\ \hline
		Water-Net {\citep{9}}          & 4.442  & -19.72  & 5.694  & 4.607  & -15.31  & 5.893 & 4.283  & -21.88 & 4.707 & \underline{4.543}  & -20.36  & 6.705 \\
		SCNet {\citep{10}}              & \underline{4.452}  & -16.85  & 5.305  & 4.597  & \underline{-11.81}  & 5.364 & \textbf{4.998}  & \underline{-17.00} & \underline{5.159} & 4.427  & -19.00  & 5.156 \\
		TACL {\citep{11}}               & 4.326  & -27.65  & 4.465  & 4.460  & -22.39  & 4.547 & 4.047  & -33.61 & 4.520 & 4.390  & -23.79  & 4.402 \\
		Spectroformer {\citep{14}}   & 4.314  & -20.33  & 5.308  & \textbf{4.881}  & -15.64  & 5.283 & 4.109  & -25.31 & 5.046 & 4.384  & -21.56  & 5.887 \\ 	\hline	
		USUIR {\citep{12}}  & 4.296  & -21.25  & 5.374  & 4.672  & -20.31  & 5.269 & 4.116  & -23.44 & 4.924 & 4.474  & -19.78  & 6.253 \\          
		PUGAN {\citep{13}}  & 4.377  & -23.03  & 5.133  & 4.542  & -17.48  & 5.366 & 4.209  & -26.55 & 5.081 & 4.487  & -21.79  & 4.990 \\    \hline        
		UVZ               & \textbf{4.619}  & \textbf{-9.933}  & \underline{6.241}  & \underline{4.863}  & \textbf{-4.661}  & \underline{6.006} & \underline{4.713}  & \textbf{-2.238} & \textbf{6.322} & \textbf{4.896}  & \textbf{5.532}   & 6.402 \\ \bottomrule
	\end{tabular}
	\label{Tab:2}
\end{table*}

Table~\ref{Tab:2} gives the quantitative metrics on the above datasets, showcasing that our method excels in most metrics, especially the UICM representing the color metric. CBM exhibits impressive NIQE metrics, signifying a natural restoration of scene details. However, its color performance heavily affected the UICM evaluation.

\subsection{Ablation Study}
\begin{figure}[ht!]
	\centering
	\includegraphics[width=8.5cm]{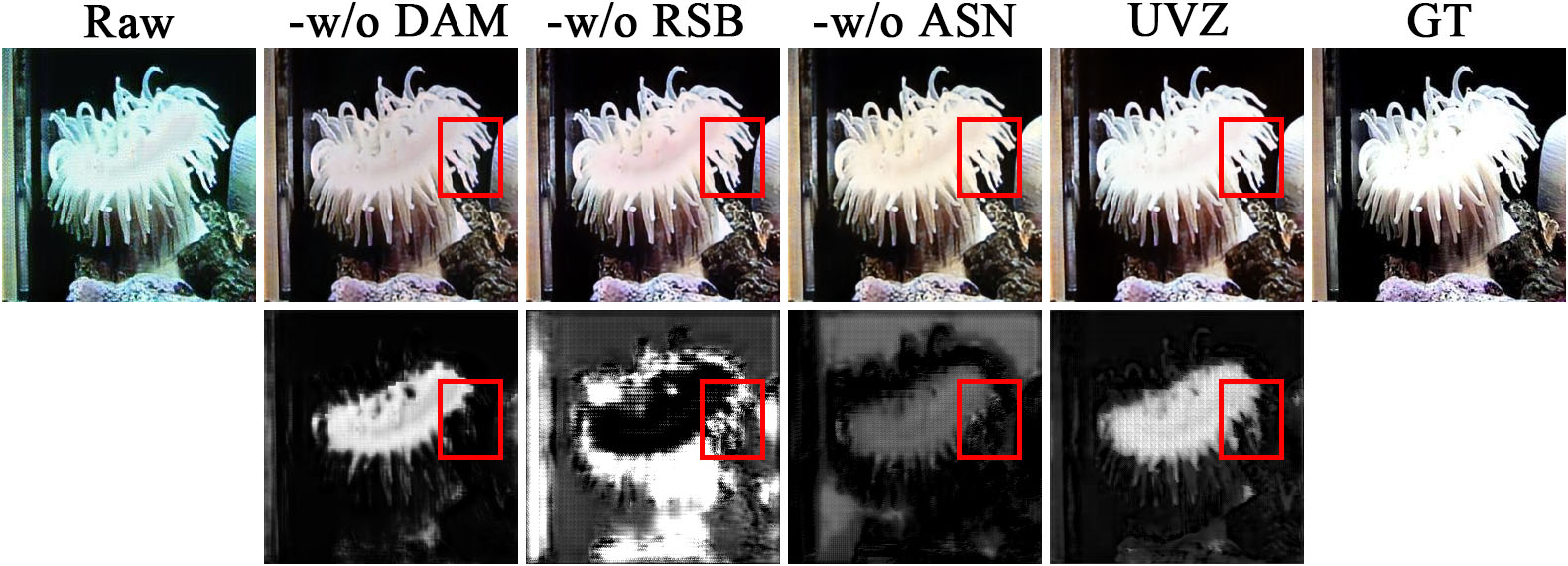}
	\caption{Ablation comparison of the first stage configurations. The top shows the raw image, enhanced images for each configuration, and GT, and the bottom shows the estimated depth maps of the four configurations.}
	\label{Fig:13}
\end{figure}

\begin{table}[h!]
	\tiny
	\centering
	\renewcommand{\arraystretch}{1.5}
	\setlength\tabcolsep{5pt}
	\caption{Metrics comparison of the first stage configurations on the CYCLE dataset (\textbf{optimal}, \underline{sub-optimal})}
	\begin{tabular}{@{}cccccc@{}}
		\toprule
		& \textbf{PSNR↑} & \textbf{MSE↓}    & \textbf{UIQM↑} & \textbf{UICM↑} & \textbf{NIQE↑} \\ \midrule
		-w/o DAM & 25.73          & 0.00438          & 4.942          & 0.909          & \underline{6.500}    \\
		-w/o   RSB & 25.99          & 0.00437          & \underline{5.016}    & \underline{3.346}    & 6.371          \\
		-w/o   ASN & \textbf{26.02} & \textbf{0.00433} & 4.883          & -0.952         & 6.410          \\
		UVZ        & \underline{26.00}    & \underline{0.00435}    & \textbf{5.044} & \textbf{4.781} & \textbf{6.586} \\ \bottomrule
	\end{tabular}
	\label{Tab:3}
\end{table}

We conducted a detailed ablation study for two stages separately. First, we removed different components in the first stage, and fixed the second stage configuration to train the entire network. The ablation configurations are as follows: (1) -w/o DAM, representing the removal of DAM in DEN, retaining only skip connections. (2) -w/o RSB, representing the removal of feature transmission from RSB in ASN. (3) -w/o ASN, representing the removal of auxiliary supervision from ASN in the training process. Figure~\ref{Fig:13} compares the effects of the above configurations. In the absence of DAM, the depth estimation of scene details lacks accuracy, and the removal of RSB introduces confusion in the predicted depth. Therefore, the enhanced images of two configurations suffer from excessive red channel enhancement. The depth image of -w/o ASN deviates from the underwater image, leading to inaccurate color enhancement. In Table~\ref{Tab:3}, although the referenced metrics of the complete network are slightly lower than that of -w/o ASN, the advantage of the no-reference metrics still proves a significant gain of ASN.

\begin{figure}[ht!]
	\centering
	\includegraphics[width=9cm]{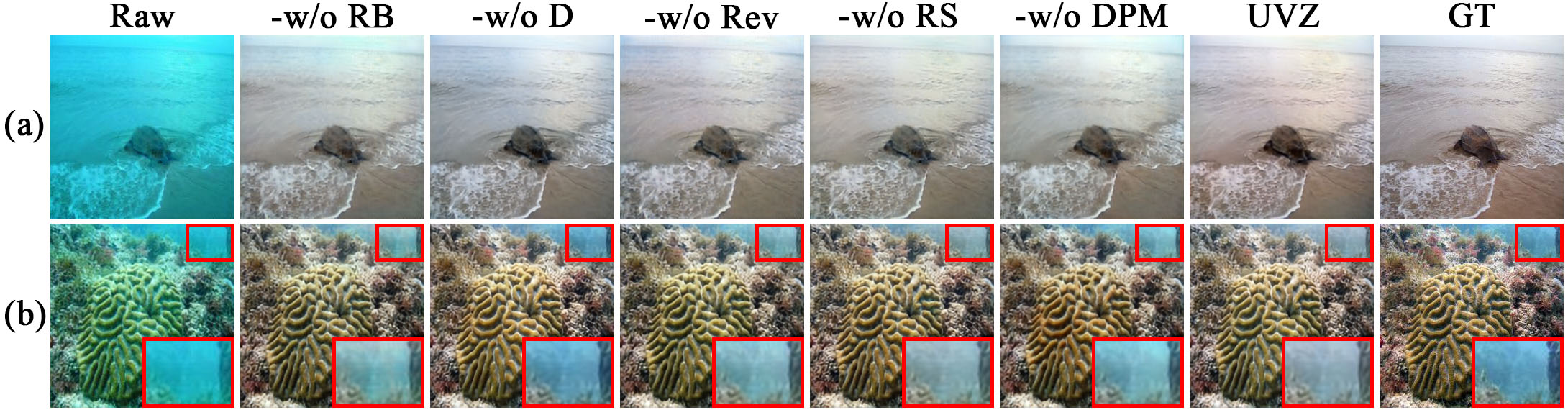}
	\caption{Ablation comparison of the second stage configurations.}
	\label{Fig:14}
\end{figure}

\begin{table}[h!]
	\tiny
	\centering
	\renewcommand{\arraystretch}{1.5}
	\setlength\tabcolsep{5pt}
	\caption{Metrics comparison of the second stage configurations on the CYCLE dataset (\textbf{optimal}, \underline{sub-optimal})}
	\begin{tabular}{@{}cccccc@{}}
		\toprule
		& \textbf{PSNR↑} & \textbf{MSE↓}    & \textbf{UIQM↑} & \textbf{UICM↑} & \textbf{NIQE↑} \\ \midrule
		-w/o RB    & 25.62          & 0.00477          & 4.966          & 0.472          & 6.292          \\			
		-w/o D     & \underline{25.76}    & 0.00448          & 5.024          & 3.613          & 6.474    \\
		-w/o Rev     & 25.73    & 0.00453          & 5.014          & 1.840          & \underline{6.504}    \\
		-w/o RS    & 25.69          & 0.00450          & \textbf{5.093} & \underline{4.733}    & 6.368          \\
		-w/o   DPM & 25.65          & \underline{0.00445}    & 5.020          & 4.478          & 6.277          \\
		\textbf{UVZ}        & \textbf{26.00} & \textbf{0.00435} & \underline{5.044}    & \textbf{4.781} & \textbf{6.586} \\ \bottomrule
	\end{tabular}
	\label{Tab:4}
\end{table}

Second, we fixed the configuration of the first stage, and removed different components in the second stage to obtain the following ablation network: (1) -w/o RB, representing the removal of all RB. (2) -w/o D, representing the removal of the depth maps in DPM, retaining only the second stage. (3) -w/o Rev, representing the removal of depth map reversal. (4) -w/o RS, representing the removal of RS in the R\(^3\)S transformation, directing the local branch only through the reshaped depth map. (5) -w/o DPM, representing the removal of the DPM, retaining only skip connections. In Figure~\ref{Fig:14}. (a), the color of the complete network is the most consistent with GT. When the network lacks RB or DPM, the enhanced image is still affected by color deviation. While the red channel enhancement of - w/o D, -w/o Rev and - w/o RS is insufficient, and the color distribution of -w/o RS is chaotic. In the enlarged far water region of Figure~\ref{Fig:14}. (b), only the -w/o RS and the complete network effectively counter the color deviation. In Table~\ref{Tab:4}, the complete network is weaker than the -w/o RS in UIQM, but it outperforms the other configured networks in the remaining four metrics.

\subsection{Extended Applications}
\begin{figure*}[ht!]
	\centering
	\includegraphics[width=12cm]{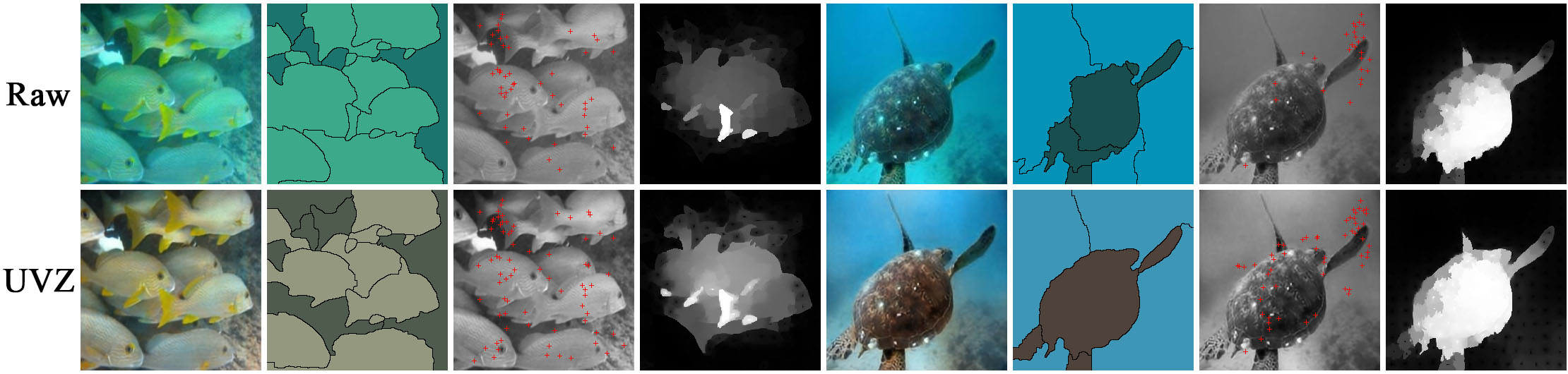}
	\caption{Effect comparison of different tasks before and after enhancement. The top shows the raw images and task results, and the bottom shows the enhanced images and task results.}
	\label{Fig:15}
\end{figure*}

\begin{figure*}[ht!]
	\centering
	\includegraphics[width=12cm]{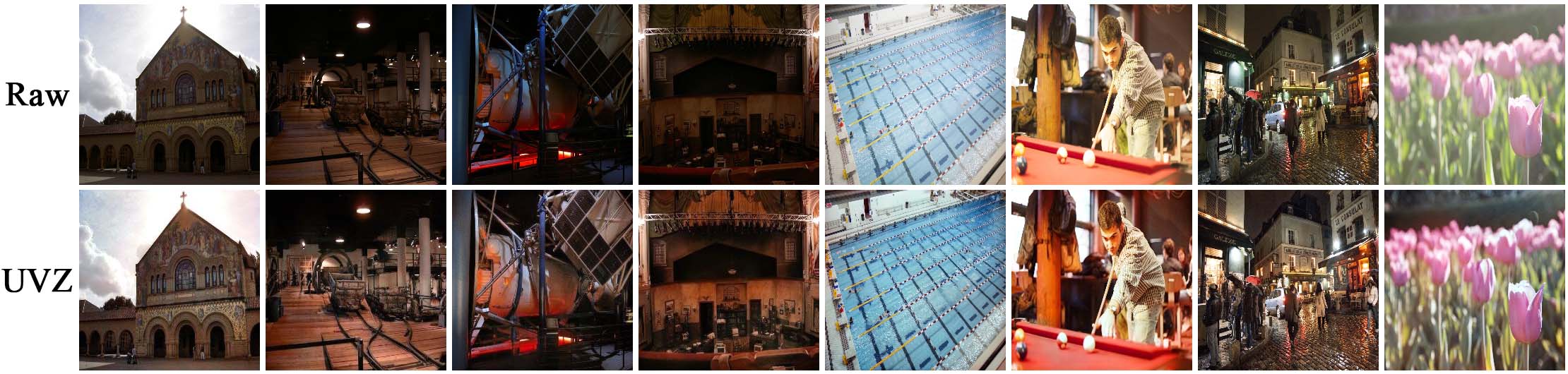}
	\caption{Enhancement effects in different lighting environments. Our method can significantly improve the visibility without parameter adjustment.}
	\label{Fig:16}
\end{figure*}

Without adjusting any parameters, we applied the model to three challenging tasks: image segmentation \citep{47}, keypoint detection \citep{48}, and saliency detection \citep{49,50}, while also promoting it in low-light \citep{51,52} and exposure \citep{53} scenarios. Figure~\ref{Fig:15} illustrates the results of the different tasks. The enhanced image segmentation exhibits more accurate details, and the number of keypoints increases from 49 and 27 to 73 and 56, respectively. Additionally, the saliency structure of the image is also more specific. In Figure~\ref{Fig:16}, the UVZ effectively enhances the brightness and details of the low-light image, and makes the colors of the exposed image more vivid. The above good applications further validate the effectiveness and generalization of UVZ.

\begin{figure*}[h!]
	\centering
	\setlength{\belowcaptionskip}{-0.3cm}
	\includegraphics[width=8cm]{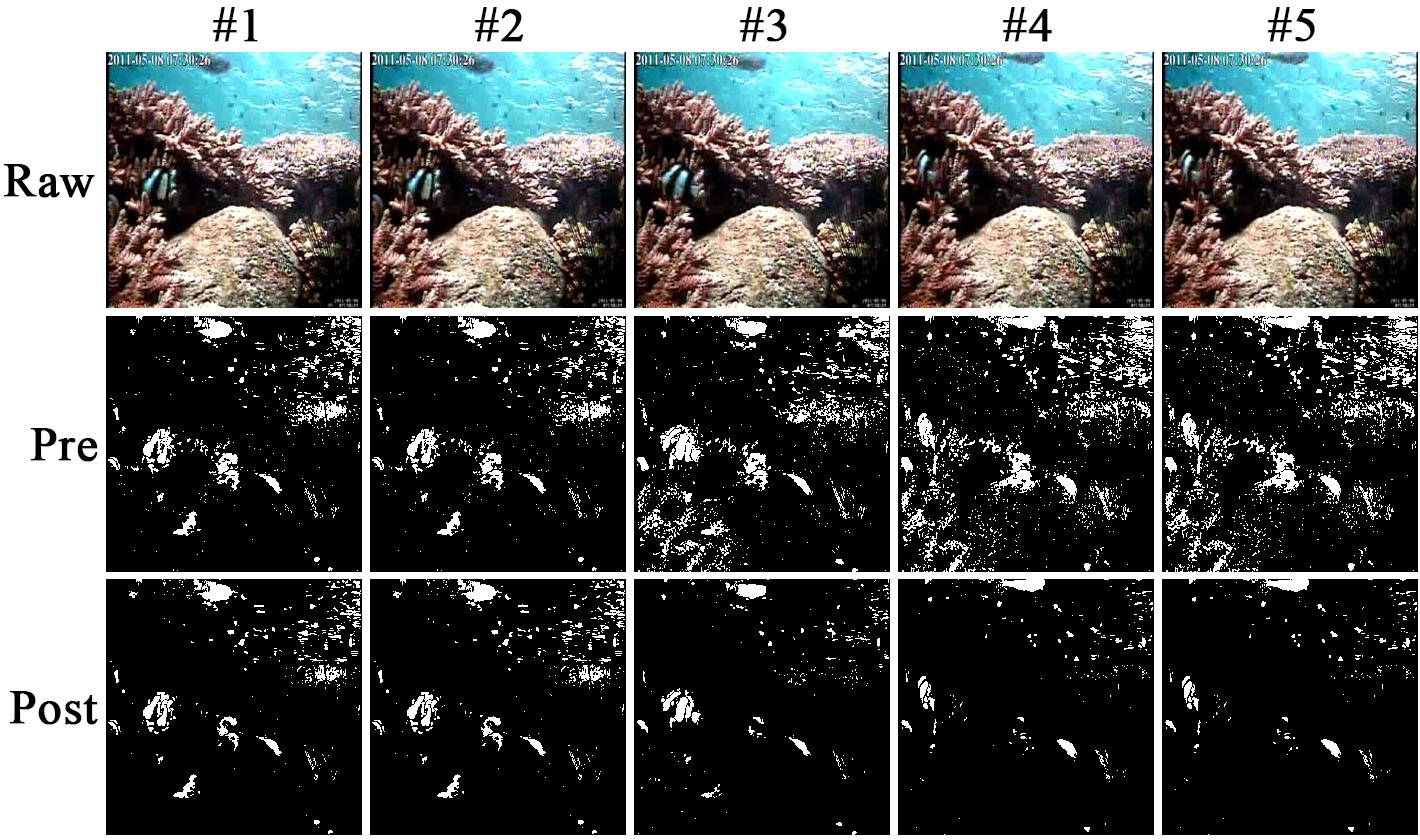}
	\caption{Comparison of pre-post detection results by the background subtraction algorithm. The top shows five consecutive frames containing notable fish movements, the middle shows the direct detection results from the algorithm, and the bottom shows the results of the algorithm after UVZ pre-processing.}
	\label{Fig:17}
\end{figure*}

To verify the gain of UVZ in real underwater task, we integrate it as a pre-processing step into the background subtraction algorithm \citep{54}, which is capable of capturing moving fish. Specifically, we selected the "Dynamic Background" sequence of video frames from UNICT \citep{55,56} dataset as a test object, given its rich variety of marine organism movements and scene changes. As shown in Figure~\ref{Fig:17}, by introducing UVZ as a pre-processing step, the algorithm's anti-interference ability against complex background is significantly improved, resulting in fish movements appearing more prominent and clearer. The result indicates that UVZ can effectively enhance the performance of fish motion detection, making it more robust in complex underwater environments.

\section {Conclusion}
This paper proposes a novel UIE framework for depth-guided perception. In DEN, we use DAM to focus on regions with severe degradation, making the predicted depth map more reasonable. To ensure the relevance of depth map to scene, ASN is introduced into the training process to generate regression image. In addition, we design a DPM to capture near-far features, with the R\(^3\)S transformation guiding the fusion of non-local and local branches. Extensive experiments on benchmark datasets demonstrate the advantages of our method and the gains of each module. Moreover, our method proves effectiveness across different visual tasks, improving visibility in challenging lighting conditions. However, UVZ exhibits limited region division in response to diverse underwater scenes, and lags behind lightweight models in speed. In future work, we will aim to optimize the model efficiency and explore multiple semantic information fusion to enhance scenario understanding.

\section*{Acknowledgments}
This work was supported by National Natural Science Foundation of China (61972064), LiaoNing Revitalization Talents Program (XLYC1806006) and the Fundamental Research Funds for the Central Universities (DUT19RC(3)012).



\bibliographystyle{elsarticle-harv} 
\bibliography{elsarticle-refs}

%
%
%
\end{document}